\documentclass[11pt,a4paper]{article}
\usepackage{authblk}

\usepackage[hyperref]{acl2020}
\usepackage{times}
\usepackage{latexsym}

\usepackage{microtype}

\aclfinalcopy % Uncomment this line for the final submission
\usepackage{graphicx}
\usepackage{amssymb}
\usepackage{mathtools}
\usepackage{multirow}

\title{RYANSQL: Recursively Applying Sketch-based Slot Fillings\\for Complex Text-to-SQL in Cross-Domain Databases}

\author[1,2]{ DongHyun Choi}
\author[1]{ Myeong Cheol Shin}
\author[1]{ EungGyun Kim}
\author[2]{ Dong Ryeol Shin}

\affil[1]{ Kakao Enterprise, Pangyo, South Korea}
\affil[ ]{\textit {\{heuristic.c,index.sh.jason.ng\}@kakaoenterprise.com}}
\affil[2]{Sungkyunkwan University, Suwon, South Korea}
\affil[ ]{\textit {drshin@skku.edu}}

\date{}

\begin{document}
\maketitle
\begin{abstract}
Text-to-SQL is the problem of converting  a user question into an SQL query, when the question and database are given. In this paper, we present a neural network approach called RYANSQL (Recursively Yielding Annotation Network for SQL) to solve complex Text-to-SQL tasks for cross-domain databases. Statement Position Code (SPC) is defined to transform a nested SQL query into a set of non-nested \texttt{SELECT} statements; a sketch-based slot filling approach is proposed to synthesize each \texttt{SELECT} statement for its corresponding SPC. Additionally, two input manipulation methods are presented to improve generation performance further. RYANSQL achieved 58.2\% accuracy on the challenging Spider benchmark, which is a 3.2\%p improvement over previous state-of-the-art approaches. At the time of writing, RYANSQL achieves the first position on the Spider leaderboard.
 
 \end{abstract}

\section{Introduction}
Text-to-SQL is the task of generating SQL queries when database and natural language user questions are given. Recently proposed neural net architectures achieved more than 80\% exact matching accuracy on the well-known Text-to-SQL benchmarks such as ATIS, GeoQuery and WikiSQL \citep{sqlnet, typesql, incsql, dong:18, sqlova, xsql}. However, those benchmarks have shortcomings of either assuming the same database across the training and test dataset(ATIS, GeoQuery) or assuming the database of a single table and restricting the complexity of SQL query to have a unitary \texttt{SELECT} statement with a single \texttt{SELECT} and \texttt{WHERE} clause (WikiSQL). 

Different from those benchmarks, the Spider benchmark proposed by \citet{Yu:18} contains complex SQL queries with cross-domain databases. Cross-domain means that the databases for the training dataset and test dataset are different; the system should predict with an unseen database as its input during testing. Also, different from another cross-domain benchmark WikiSQL, the SQL queries in Spider benchmark contain nested queries with multiple \texttt{JOIN}ed tables, and clauses like \texttt{ORDERBY}, \texttt{GROUPBY}, and \texttt{HAVING}. \citet{Yu:18} showed that the state-of-the-art systems for the previous benchmarks do not perform well on the Spider dataset.

In this paper, we propose a novel network architecture called RYANSQL (Recursively Yielding Annotation Network for SQL) to handle such complex, cross-domain Text-to-SQL problem. The proposed approach generates nested queries by recursively yielding its component \texttt{SELECT} statements. A sketch-based slot filling approach is proposed to predict each \texttt{SELECT} statement. In addition, two simple but effective input manipulation methods are proposed to improve the overall system performance. The proposed system improves the previous state-of-the-art system by 3.2\%p in terms of the test set exact matching accuracy, using BERT \citep{Devlin:19}. Our contributions are summarized as follows.

\begin{itemize}
\item{We propose a detailed sketch for the complex \texttt{SELECT} statements, along with a network architecture to fill the slots. }
\item{Statement Position Code (SPC) is introduced to recursively predict nested queries with sketch-based slot filling algorithm.}
\item{We suggest two simple input manipulation methods to improve performance further. Those methods are easy to apply, and they improve the overall system performance significantly.}
\end{itemize}

\section{Related Works}
\label{sec:related}
Most recent works on Text-to-SQL task used encoder-decoder model. Those works could be classified into three main categories, based on their decoder outputs. Direct Sequence-to-Sequence approaches \citep{dong:16,zhong:17} generate SQL query tokens as their decoder outputs; due to the risk of generating grammatically incorrect SQL queries, the Direct Sequence-to-Sequence approaches are rarely used in recent works. 

Grammar-based approaches generate a sequence of grammar rules and apply the generated rules sequentially to get the resultant SQL query. \citet{incsql} defines a structural representation of an SQL query and a set of parse actions to handle the WikiSQL dataset. \citet{irnet} defines SemQL queries, which is an abstraction of SQL query in tree form, along with a set of grammar rules to synthesize SemQL queries; Synthesizing SQL query from a SemQL tree structure is straightforward. \citet{gnn} focused on the DB constraints selection problem during the grammar decoding process; they applied global reasoning between question words and database columns/tables.

Sketch-based slot-filling approaches, firstly proposed by \citet{sqlnet} to handle the WikiSQL dataset, define a sketch with slots for SQL queries, and the decoder classifies values for those slots. Sketch-based approaches \citet{sqlova} and \citet{xsql} achieved state-of-the-art performances on the WikiSQL dataset, with the aid of BERT \citep{Devlin:19}. However, sketch-based approach on more complex Spider benchmark \citep{rcsql} showed relatively low performance compared to the grammar-based approaches. There are two major reasons: (1) It is hard to define a sketch for Spider queries since the allowed syntax of the Spider SQL queries is far more complicated than that of the WikiSQL queries. (2) Since the sketch-based approaches fill values for the predefined slots, the approaches have difficulties in predicting the nested queries.

In this paper, a sketch-based slot filling approach is proposed to solve the complex Text-to-SQL problem. A detailed sketch for complex \texttt{SELECT} statements is newly proposed, along with Statement Position Code (SPC) to effectively predict for the nested queries.

\section{Task Definition}
\label{sec:task}
The Text-to-SQL task considered in this paper is defined as follows: Given a question with $n$ tokens $Q = \{w^Q_1, ..., w^Q_n\}$ and a DB schema with $t$ tables and $f$ foreign key relations $D=\{T_1, ...,T_t, F_1, ..., F_f\}$, find $S$, the SQL translation of $Q$. Each table $T_i$ consists of a table name with $t_i$ words $\{w^{T_i}_1, ..., w^{T_i}_{t_i}\}$, and a set of columns $\{C_j, ..., C_k\}$. Each column $C_j$ consists of a column name $\{w^{C_j}_1, ..., w^{C_j}_{c_j}\}$, and a marker to check if the column is a primary key.  

For an SQL query $S$ we define a non-nested form of $S$, $N(S)=\{(P_1, S_1), ..., (P_l, S_l)\}$. In the definition, $P_i$ is the $i$-th SPC, and $S_i$ is its corresponding \texttt{SELECT} statement. Table \ref{tbl:ex} shows an example of natural language query $Q$, SQL translation $S$ and its non-nested form $N(S)$.

\begin{table}
\centering
\begin{tabular}{|c|c|l|} \hline
\multirow{2}{*}{\textbf{Q}} &\multicolumn{2}{l|}{ Find the name of airports which do not  } \\
&\multicolumn{2}{l|}{have any flight in and out.}\\ \hline
\multirow{5}{*}{\textbf{S}} &\multicolumn{2}{l|}{ \texttt{SELECT} AirportName \texttt{FROM} Airports } \\
&\multicolumn{2}{l|}{ \texttt{WHERE} AirportCode \texttt{NOT} \texttt{IN} } \\
&\multicolumn{2}{l|}{ (\texttt{SELECT} SourceAirport \texttt{FROM} Flights }\\
&\multicolumn{2}{l|}{ \texttt{UNION} } \\
&\multicolumn{2}{l|}{ \texttt{SELECT} DestAirport \texttt{FROM} Flights) }\\ \hline
\multirow{8}{*}{\textbf{N(S)}}&$P_1$&[ \texttt{NONE} ] \\  \cline{2-3}
&\multirow{3}{*}{$S_1$}& \texttt{SELECT} AirportName \\
 & & \texttt{FROM} Airports\\ 
  & & \texttt{WHERE} AirportCode \texttt{NOT} \texttt{IN} $S_2$ \\ \cline{2-3}
&$P_2$& [ \texttt{WHERE} ] \\ \cline{2-3}
&\multirow{2}{*}{$S_2$}&\texttt{SELECT} SourceAirport \\ 
& & \texttt{FROM} Flights \texttt{UNION} $S_3$\\ \cline{2-3}
&$P_3$&[ \texttt{WHERE}, \texttt{UNION} ] \\ \cline{2-3}
&$S_3$&\texttt{SELECT} DestAirport \texttt{FROM} Flights \\ \hline
\end{tabular}
\caption{Example of a user question $Q$, SQL translation $S$, and its non-nested form $N(S)$.}
\label{tbl:ex}
\end{table}

\begin{figure*}
\centering
\includegraphics[width=\textwidth]{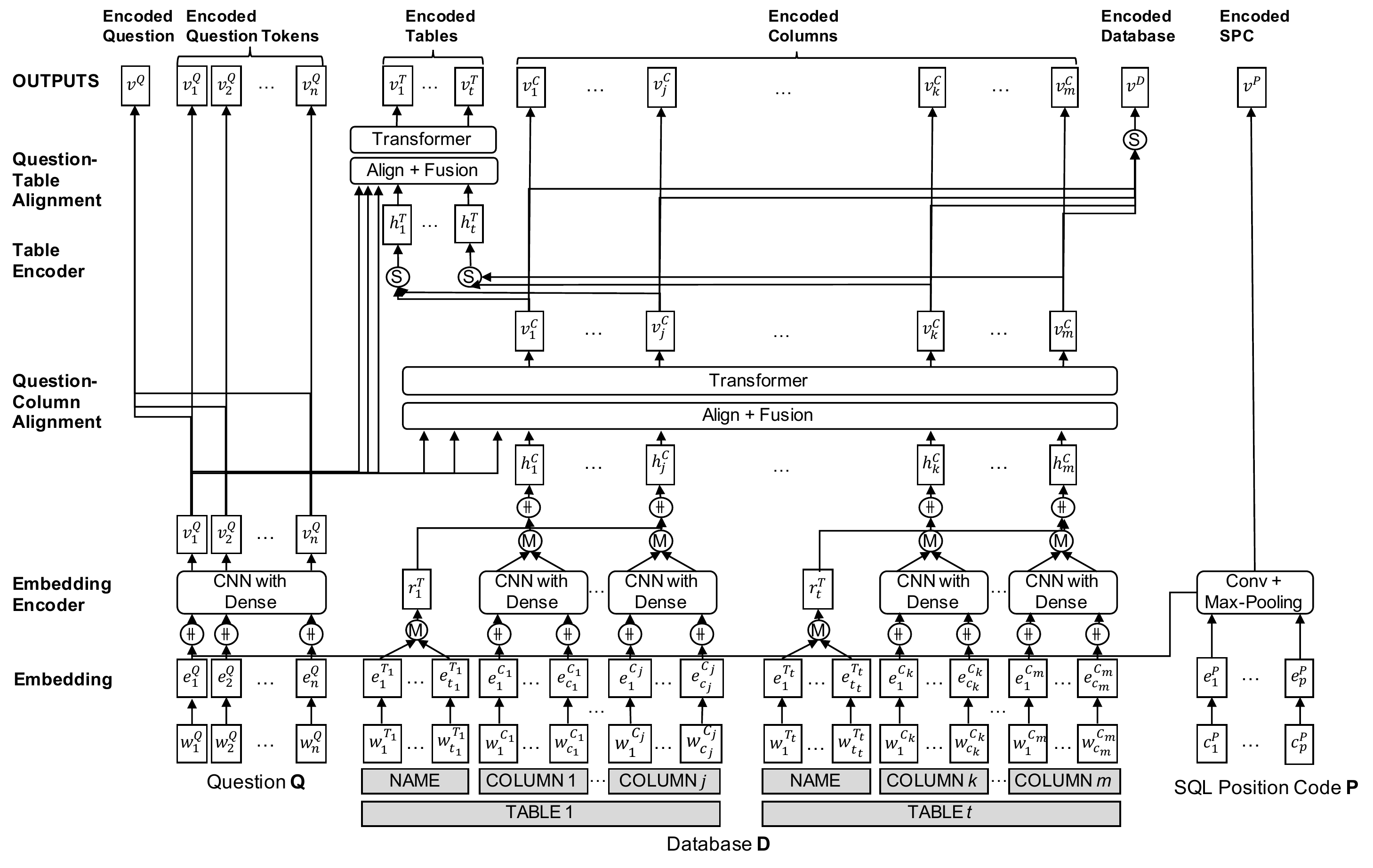}
\caption{Network architecture of the proposed input encoder. \raisebox{-0.5ex}{\includegraphics[height=3.3mm]{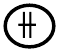}} represents vector concatenation, \raisebox{-0.5ex}{\includegraphics[height=3.3mm]{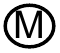}} represents max-pooling and \raisebox{-0.5ex}{\includegraphics[height=3.3mm]{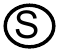}} represents self-attention.}
\label{fig:encode}
\end{figure*}

Each SPC $P$ could be considered as a sequence of $p$ position code elements, $P = [ c^P_1, ..., c^P_p ]$. The possible set of position code elements is $\{$\texttt{NONE}, \texttt{UNION}, \texttt{INTERSECT}, \texttt{EXCEPT}, \texttt{WHERE}, \texttt{HAVING}, \texttt{PARALLEL}$\}$. \texttt{NONE} represents the outermost statement, while \texttt{PARALLEL} means the parallel elements inside a single clause like the second element of the \texttt{WHERE} clause. Other position code elements represent corresponding SQL clauses.

Since it is straightforward to construct $S$ from $N(S)$, the goal of the proposed system is to construct $N(S)$ for the given $Q$ and $D$. To achieve the goal, the proposed system first sets the initial SPC $P_1=[\texttt{NONE}]$, and predicts its corresponding \texttt{SELECT} statement and nested SPCs. The system recursively finds out the corresponding \texttt{SELECT} statements for remaining SPCs, until every SPC has its corresponding \texttt{SELECT} statement. 

\section{Generating a \texttt{SELECT} Statement}
\label{sec:genselect}

In this section, the method to create the \texttt{SELECT} statement for the given question $Q$, database $D$, and SPC $P$ is described. Section \ref{subsec:encoder} describes the input encoder; the sketch-based slot-filling decoder is described in section \ref{subsec:decoder}.

\subsection{Input Encoder}
\label{subsec:encoder}

Figure \ref{fig:encode} shows the overall network architecture. The input encoder consists of five layers: Embedding layer, Embedding Encoder layer, Question-Column Alignment layer, Table Encoder layer, and Question-Table Alignment layer. 

\paragraph{Embedding.} To get the embedding vector for a word $w$ in question, table names or column names, its word embedding and character embedding are concatenated. The word embedding is initialized with $d_1 = 300$-dimensional pre-trained GloVe \citep{pennington14} word vectors, and is fixed during training. For character embedding, each character is represented as a trainable vector of dimension $d_2 = 50$, and we take maximum value of each dimension of component characters to get the fixed-length vector. The two vectors are then concatenated to get the embedding vector $e_w \in \mathbb{R}^{d_1 + d_2}$.  One layer highway network \citep{highway} is applied on top of this representation. For SPC $P$, each code element  $c$ is represented as a trainable vector of dimension $d_p = 100$. 

\paragraph{Embedding Encoder.} One dimensional convolution layer with kernel size 3 is applied on SPC element embedding vectors $\{e^P_1, ..., e^P_p\}$. Max-pooling is applied on the output to get the SPC vector $v^P \in \mathbb{R}^{p \times d_p}$. $v^P$ is then concatenated to each question and column word embedding vector. 

CNN with Dense Connection proposed in \citet{Yoon18} is applied to encode each word of question and columns to capture local information. Parameters are shared across the question and columns. For each column, a max-pooling layer is followed; outputs are concatenated with their table name representations and projected to dimension $d$. Layer outputs are encoded question words $V^Q=\{v^Q_1, v^Q_2, ..., v^Q_n\} \in \mathbb{R}^{n \times d}$, and hidden column vectors $H^C = \{h^C_1, ..., h^C_m\} \in \mathbb{R}^{m \times d}$.

\paragraph{Question-Column Alignment.} In this layer, the column vectors are modeled with contextual information of the question by attending question tokens with their corresponding columns. Scaled dot-product attention \citep{transformer} is used to align question tokens with column vectors:

\begin{equation}
\begin{aligned}
A_{QtoC} &=\text{softmax}( \frac{H^C \cdot (V^Q)^\intercal}{\sqrt{d}} ) \cdot V^Q \\
\end{aligned}
\end{equation}

Where each $i$-th row of $A_{QtoC} \in \mathbb{R}^{m \times d}$ is an attended question vector of the $i$-th column.

The heuristic fusion function $fusion(x,y)$, proposed in \citet{mnemonic}, is applied to merge $A_{QtoC}$ with $H^C$:
\begin{align}
\tilde{x} &= \text{relu}(W_r[x;y;x \circ y; x - y ]) \nonumber \\ 
g &= \sigma (W_g[x;y;x \circ y; x - y ]) \nonumber \\ 
fusion(x,y) &= g \circ \tilde{x} + ( 1 -g ) \circ x \nonumber \\
F^C &= fusion( A_{QtoC}, H^C ) 
\end{align}

Where $W_r$ and $W_g$ are trainable variables, $\sigma$ denotes the sigmoid function, $\circ$  denotes element-wise multiplication and $F^C \in \mathbb{R}^{m \times d}$ is fused column matrix. A transformer layer \citep{transformer} is applied on top of $F^C$ to capture contextual column information. Layer outputs are the encoded column vectors $V^C=\{v^C_1, ..., v^C_m\} \in \mathbb{R}^{m \times d}$.

\paragraph{Table Encoder.} Column vectors belonging to each table are integrated to get the encoded table vector. For a matrix $M \in \mathbb{R}^{n \times d}$, self-attention function $f_s(M) \in \mathbb{R}^{1 \times d}$ is defined as follows:

\begin{equation}
\begin{aligned}
f_s(M) &= \text{softmax}( W_2 \tanh(W_1M^\intercal) )M
\end{aligned}
\end{equation}

Where $W_1 \in \mathbb{R}^{d \times d}$, $W_2 \in \mathbb{R}^{1 \times d}$ are trainable parameters. Then, for table $t$ with columns $\{C_j, ..., C_k\}$, the hidden table vector $h^T_t$ is calculated as follows:

\begin{align}
h^T_t &= f_s([v^C_j;...;v^C_k])
\end{align}

Layer outputs are the hidden table vectors $H^T=\{h^T_1, h^T_2, ..., h^T_t\} \in \mathbb{R}^{t \times d}$.

\paragraph{Question-Table Alignment.} In this layer, the same network architecture as the Question-Column alignment layer is used to model the table vectors with contextual information of the question. Layer outputs are the encoded table vectors $V^T = \{v^T_1, v^T_2, ..., v^T_t\} \in \mathbb{R} ^ {t \times d}$.

\paragraph{Output.} Final outputs of the input encoder are: (1) Encoded question words $V^Q = \{v^Q_1, ..., v^Q_n\} \in \mathbb{R}^{n \times d}$, (2) Encoded columns $V^C = \{v^C_1, ..., v^C_m\} \in \mathbb{R}^{m \times d}$, (3) Encoded tables $V^T = \{v^T_1, ..., v^T_t\} \in \mathbb{R}^{t \times d}$, and (4) Encoded SPC $v^P \in \mathbb{R}^{d_p}$. Additionally, (5) Encoded question $v^Q = f_s(V^Q)$ and (6) Encoded DB schema $v^D = f_s(V^C) \in \mathbb{R}^d$ are calculated for later use.

\subsubsection{BERT-based Input Encoder}
Inspired by the work of \citet{sqlova} and \citet{irnet}, BERT \citep{Devlin:19} is considered as another version of input encoder. The input to BERT is constructed by concatenating question words, SPC elements and column words as follows: \texttt{[CLS]}, $w^Q_i$, \texttt{[SEP]}, $c^P_j$, \texttt{[SEP]}, $w^{C_1}_k$, \texttt{[SEP]}, ..., $w^{C_m}_l$, \texttt{[SEP]}. 

Hidden states of the last layer are retrieved to form $V^Q$ and $V^C$; for $V^C$, the state of each column's last word is taken to represent encoded column vector. Each table vector $v^T_j$ is calculated as a self-attended vector of its containing columns; $v^Q$, $v^D$, and $v^P$ are calculated as the same.

\subsection{Sketch-based Slot-Filling Decoder} 
\label{subsec:decoder}

\begin{table*}[hbt]
\centering
\begin{tabular}{|l|l|} \hline
\textbf{CLAUSE}&\textbf{SKETCH} \\ \hline
\texttt{FROM}& $(\texttt{\$TBL})^+$\\ \hline
\texttt{SELECT}& \texttt{\$DIST} ( \texttt{\$AGG} ( \texttt{\$DIST$_1$} \texttt{\$AGG$_1$} \texttt{\$COL$_1$} \texttt{\$ARI} \texttt{\$DIST$_2$} \texttt{\$AGG$_2$} \texttt{\$COL$_2$} ) ) $^+$  \\ \hline
\texttt{ORDERBY}&  ( ( \texttt{\$DIST$_1$} \texttt{\$AGG$_1$} \texttt{\$COL$_1$} \texttt{\$ARI} \texttt{\$DIST$_2$} \texttt{\$AGG$_2$} \texttt{\$COL$_2$} ) \texttt{\$ORD} ) $^*$ \\ \hline
\texttt{GROUPBY} & ( \texttt{\$COL} )$^*$ \\ \hline
\texttt{LIMIT} & \texttt{\$NUM} \\ \hline
\texttt{WHERE} &  ( \texttt{\$CONJ} ( \texttt{\$DIST$_1$} \texttt{\$AGG$_1$} \texttt{\$COL$_1$} \texttt{\$ARI} \texttt{\$DIST$_2$} \texttt{\$AGG$_2$} \texttt{\$COL$_2$} ) \\
\texttt{HAVING} & ~~\texttt{\$NOT}  \texttt{\$COND} \texttt{\$VAL}$_1$$|$\texttt{\$SEL}$_1$  \texttt{\$VAL}$_2$$|$\texttt{\$SEL}$_2$  )$^*$ \\ \hline
\texttt{INTERSECT} \ & \\
\texttt{UNION} \ & \texttt{\$SEL} \\
\texttt{EXCEPT} \ & \\ \hline
\end{tabular}
\caption{Proposed sketch for a \texttt{SELECT} statement. \texttt{\$TBL} and \texttt{\$COL} represent a table and a column, respectively. \texttt{\$AGG} is one of \{\textbf{none}, \textbf{max}, \textbf{min}, \textbf{count}, \textbf{sum}, \textbf{avg}\}, \texttt{\$ARI} is one of the arithmetic operators \{\textbf{none}, \textbf{-}, \textbf{+}, \textbf{ *}, \textbf{/} \}, and \texttt{\$COND} is one of the conditional operators \{\textbf{between}, \textbf{=}, \textbf{\textgreater}, \textbf{\textless}, \textbf{\textgreater=}, \textbf{\textless=}, \textbf{!=}, \textbf{in}, \textbf{like}, \textbf{is}, \textbf{exists}\}. \texttt{\$DIST} and \texttt{\$NOT} are boolean variables representing the existence of keywords \texttt{DISTINCT} and \texttt{NOT}, respectively. \texttt{\$ORD} is a binary value for keywords \texttt{ASC}/\texttt{DESC}, and \texttt{\$CONJ} is one of conjunctions \{\texttt{AND}, \texttt{OR}\}. \texttt{\$VAL} is the value for \texttt{WHERE/HAVING} condition; \texttt{\$SEL} represents the slot for another \texttt{SELECT} statement.  }
\label{tbl:sketch}
\end{table*}

Table \ref{tbl:sketch} shows the proposed sketch for a \texttt{SELECT} statement. The sketch-based slot-filling decoder predicts values for slots of the proposed sketch.  

\paragraph{Classifying Base Structure.}  By the term \textit{base structure} of a \texttt{SELECT} statement, we refer to the existence of its component clauses and the number of conditions for each clause. We first combine the encoded vectors $v^Q$, $v^D$ and $v^P$ to get the statement encoding vector $v^S$, as follows:

\begin{align}
hc( x, y ) &= concat( x, y, |x-y|, x \circ y) \label{eq:hc} \\
v^S &= W \cdot concat( hc( v^Q, v^D ), v^P )
\end{align}

Where $W \in \mathbb{R} ^ {d \times (4d + d_p)}$ is a trainable parameter, and function $hc(x,y)$ is the concatenation function for heuristic matching method proposed in \citet{heuristicmatching}.

Eleven values $b_g, b_o, b_l, b_w, b_h, n_g, n_o, n_s, n_w$, $n_h$ and $c_\texttt{IUEN}$ are classified by applying two fully-connected layers on $v^S$. Binary values $b_g, b_o, b_l, b_w, b_h$ represent the existence of \texttt{GROUPBY}, \texttt{ORDERBY}, \texttt{LIMIT}, \texttt{WHERE} and \texttt{HAVING}, respectively. Note that \texttt{FROM} and \texttt{SELECT} clauses must exist to form a valid \texttt{SELECT} statement. $n_g, n_o, n_s, n_w, n_h$ represent the number of conditions in \texttt{GROUPBY}, \texttt{ORDERBY}, \texttt{SELECT}, \texttt{WHERE} and \texttt{HAVING} clause, respectively. The maximal number of conditions $N_g = 3$, $N_o = 3$, $N_s=6$, $N_w=4$, and $N_h=2$ are defined for  \texttt{GROUPBY}, \texttt{ORDERBY}, \texttt{SELECT}, \texttt{WHERE} and \texttt{HAVING} clauses, to solve the problem as $n$-way classification problem. The values of maximal condition numbers are chosen to cover all the training cases.

Finally, $c_\texttt{IUEN}$ represents the existence of one of \texttt{INTERSECT}, \texttt{UNION} or \texttt{EXCEPT}, or \texttt{NONE} if no such clause exists. If the value of $c_\texttt{IUEN}$ is one of \texttt{INTERSECT}, \texttt{UNION} or \texttt{EXCEPT}, the corresponding SPC is created, and the \texttt{SELECT} statement for that SPC is generated recursively.

\paragraph{\texttt{FROM} clause.} A list of \texttt{\$TBL}s should be decided to predict the \texttt{FROM} clause. For each table $i$, $P_\textbf{fromtbl}(i | Q,D,P)$, the probability that table $i$ is included in the \texttt{FROM} clause, is calculated:
\begin{align}
c_i &= concat( v^T_i, v^Q, v^D, v^P ) \nonumber \\  
s_i &= W_2 \tanh( W_1 \cdot c_i )  \label{eq:tbl}  \\
P_\textbf{fromtbl}(i|Q,D,P) &=  \sigma ( s )_i \nonumber
\end{align} 

Where $W_1$, $W_2$ are trainable variables, $s=[s_1, ..., s_t] \in \mathbb{R}^t$ and $\sigma$ denotes the sigmoid function. From now on, we omit the notations $Q$, $D$ and $P$ for the sake of simplicity.

Top $n_t$ tables with the highest $P_\textbf{fromtbl}(i)$ values are chosen. We set an upper bound $N_t=6$ on possible number of tables. The formula to get $P_\textbf{\#tbl}(n_t)$ for each possible $n_t$ is:

\begin{equation}
\begin{aligned}
v^{T^\prime} &= \text{softmax}(s) \cdot V^T \\
P_\textbf{\#tbl}(n_t) &= \text{softmax}(full_2(v^{T^\prime}) )
\end{aligned}
\end{equation}

In the equation, $full_2$ means the application of two fully-connected layers, and table score vector $s$ is from equation \ref{eq:tbl}.

\paragraph{\texttt{SELECT} clause.}  The decoder first generates $N_s$ conditions to predict the \texttt{SELECT}. Since each condition depends on different parts of $Q$, we calculate attended question vector for each condition:

\begin{equation}
\begin{aligned}
A^Q_\textbf{sel}&=W_3\tanh(V^Q \cdot W_1 + v^P \cdot W_2 )^\intercal \\
V^Q_\textbf{sel} &= \text{softmax}(A^Q_\textbf{sel}) \cdot V^Q 
\end{aligned}
\label{eq:clatt}
\end{equation}

While $W_1, W_2 \in \mathbb{R}^{d \times d}$, $W_3 \in \mathbb{R}^{N_s \times d }$ are trainable parameters, and $V_\textbf{sel}^Q \in \mathbb{R}^{N_s \times d}$ is the matrix of attended question vectors for $N_s$ conditions. $v^P$ is tiled to match the row of $V^Q$.

For $m$ columns and $N_s$ conditions, $P_\textbf{sel\_col1} \in \mathbb{R}^{N_s \times m}$, the probability matrix for each column to fill the slot \texttt{\$COL$_1$} of each condition, is calculated as follows:

\begin{align}
A^C_\textbf{sel}[i] &= W_6\tanh( V^Q_\textbf{sel}[i] \cdot W_4 + V^C \cdot W_5 )^\intercal  \nonumber \\
P_\textbf{sel\_col1}[i] &= \text{softmax}( A^C_\textbf{sel}[i] ) \label{eq:col1}
\end{align}

Where $W_4, W_5 \in \mathbb{R}^{d \times d}$ and $W_6 \in \mathbb{R}^{1 \times d}$ are trainable parameters. The notation $M[i]$ is used to represent the $i$-th row of matrix $M$.

The attended question vectors are then updated with selected column information to get the updated question vector  $U^Q_\textbf{sel\_col1} \in \mathbb{R}^{N_s \times d}$:

\begin{equation}
\begin{aligned}
U^C_\textbf{sel\_col1}[i] &= P_\textbf{sel\_col1}[i] \cdot V^C \\
U^Q_\textbf{sel\_col1}[i] &= W_7 \cdot hc( V^Q_\textbf{sel} [i], U^C_\textbf{sel\_col1}[i]  ) 
\end{aligned}
\label{eq:updatedq}
\end{equation}

Where $W_7$ is a trainable variable, and $hc(x,y)$ is defined in equation \ref{eq:hc}. The probabilities for \texttt{\$DIST$_1$}, \texttt{\$AGG$_1$}, \texttt{\$ARI} and \texttt{\$AGG} are calculated by applying a fully connected layer on $U^Q_\textbf{sel\_col1}[i]$.

Equation \ref{eq:col1} is reused to calculate $P_\textbf{sel\_col2}$, with $V^Q_\textbf{sel}[i]$ replaced by $U^Q_\textbf{sel\_col1}[i]$; then $U^Q_\textbf{sel\_col2}$ is retrieved in the same way as equation \ref{eq:updatedq}, and the probabilities of \texttt{\$DIST$_2$} and \texttt{\$AGG$_2$} are calculated in the same way as \texttt{\$DIST$_1$} and \texttt{\$AGG$_1$}. Finally, the \texttt{\$DIST} slot, \texttt{DISTINCT} marker for overall \texttt{SELECT} clause, is calculated by applying a fully-connected layer on $v^S$.

Once all the slots are filled for $N_s$ conditions, the decoder retrieves the first $n_s$ conditions to predict the \texttt{SELECT} clause. That is possible since the CNN with Dense Connection used for question encoding \citep{Yoon18} captures relative position information. In combine with the SQL consistency protocol of the Spider benchmark \citep{Yu:18}, we expect the conditions are ordered in the same way as they are presented in $Q$. For the datasets without such consistency protocol, the proposed slot filling method could easily be changed to an LSTM-based model, as shown in \citet{sqlnet}.  

\paragraph{\texttt{ORDERBY} clause.} The same network structure as a \texttt{SELECT} clause is applied. The only difference is the prediction for \texttt{\$ORD} slot; this could be done by applying a fully connected layer on $U^Q_\textbf{ob\_col1}$, which is the correspondence of $U^Q_\textbf{sel\_col1}$.

\paragraph{\texttt{GROUPBY} clause.} The same network structure  as a \texttt{SELECT} clause is applied. For the \texttt{GROUPBY} case, retrieving only the values of $P_\textbf{gb\_col1}$ is enough to fill the necessary slots.

\paragraph{\texttt{LIMIT} clause.} Questions do not contain the \texttt{\$NUM} slot value for \texttt{LIMIT} clauses explicitly in many cases, if the questions are for the top-1 result (Example: ``Show the name and the release year of the song by \textbf{the youngest} singer"). Thus, the \texttt{LIMIT} decoder first determines if the given $Q$ requests for the top-1 result. If so, the decoder sets the \texttt{\$NUM} value to 1; otherwise, it tries to find the specific token for \texttt{\$NUM} among the tokens of $Q$ using pointer network \citep{pointer}. \texttt{LIMIT} top-1 probability $P_{\textbf{limit\_top1}}$ is retrieved by applying a fully-connected layer on $v^S$. $P^Q_{\textbf{limit\_num}}[i]$, the probability of $i$-th question token for \texttt{\$NUM} slot value, is calculated as:

\begin{align}
A^Q_{\textbf{limit\_num}} &= W_3\tanh(V^Q \cdot W_1 + v^P \cdot W_2 )^\intercal \nonumber \\
P^Q_{\textbf{limit\_num}}[i] &= \text{softmax} (A^Q_{\textbf{limit\_num}})_i
\end{align}

$W_1$, $W_2 \in \mathbb{R}^{d \times d}$, $W_3 \in \mathbb{R}^{1 \times d}$ are trainable parameters.

\paragraph{\texttt{WHERE} clause.} The same network structure as a \texttt{SELECT} clause is applied to get the attended question vectors $V^Q_\textbf{wh} \in \mathbb{R}^{N_w \times d}$, and probabilities for \texttt{\$COL$_1$}, \texttt{\$COL$_2$}, \texttt{\$DIST$_1$}, \texttt{\$DIST$_2$}, \texttt{\$AGG$_1$}, \texttt{\$AGG$_2$} and \texttt{\$ARI}. Besides, a fully-connected layer is applied on $U^Q_\textbf{wh\_col1}$ to get the probabilities for \texttt{\$CONJ}, \texttt{\$NOT} and \texttt{\$COND}.

 A fully-connected layer is applied on $U^Q_\textbf{wh\_col1}$ and $U^Q_\textbf{wh\_col2}$ to determine if the condition value for each column is another nested \texttt{SELECT} statement or not. If the value is determined as a nested \texttt{SELECT} statement, the corresponding SPC is generated, and the \texttt{SELECT} statement for the SPC is predicted recursively. If not, the pointer network is used to get the start and end position of the value span from question tokens.

\paragraph{\texttt{HAVING} clause.} The same network structure as a \texttt{WHERE} clause is applied.

\section{Two Input Manipulation Methods}
\label{sec:data}

\begin{table}
\centering
\begin{tabular}{|l|} \hline
\textbf{Q:} What are the papers of Liwen Xiong in 2015? \\
\textbf{SQL:} \\
\texttt{SELECT DISTINCT} t3.paperid \\
~~\texttt{FROM} writes \texttt{AS} t2 \\
~~~~\texttt{JOIN} author \texttt{AS} t1 \texttt{ON} t2.authorid  =  t1.authorid \\
~~~~\texttt{JOIN} paper \texttt{AS} t3 \texttt{ON} t2.paperid  =  t3.paperid  \\
~~\texttt{WHERE} t1.authorname  =  "Liwen Xiong"  \\
~~~~\texttt{AND} t3.year  =  2015; \\ \hline
\end{tabular}
\caption{SQL statement with link table. Table \textbf{writes} is not explicitly mentioned in $Q$, but it is used in the \texttt{JOIN} statement to link between tables \textbf{author} and \textbf{paper}.}
\label{tbl:linktbl}
\end{table}

In this section, we introduce two input manipulation methods to improve the performance of our proposed system further.

\subsection{\texttt{JOIN} Table Filtering}
In a \texttt{FROM} clause, some tables may be used only to make ``link" between other tables; Table \ref{tbl:linktbl} shows such an example. Those ``link" tables are necessary to create the proper \texttt{SELECT} statement, but they work as noise for Question-Table alignment since they do not have the corresponding tokens in $Q$. Thus, we discard those tables from \texttt{FROM} clauses during training; while inferencing, the link tables are easily recovered using foreign key relations.

\subsection{Supplemented Column Names}

\begin{table}
\centering
\begin{tabular}{|l|l|l|}\hline
\textbf{Table}&\textbf{Column}&\textbf{SCN} \\ \hline
\multirow{2}{*}{tv channel} & id & tv channel id \\ \cline{2-3}
& series name & tv channel series name \\ \hline
tv series & id & tv series id \\ \hline
cartoon & id & cartoon id \\ \hline
\end{tabular}
\caption{Examples of supplemented column names. \textbf{SCN} represents Supplemented Column Name.}
\label{tbl:scn}
\end{table}

We supplement the column names with its table names to distinguish between columns with the same name but belonging to different tables and representing different entities. Table names are concatenated in front of their belonging column names to form SCNs, but if the stemmed form of a table name is wholly included in a stemmed form of the column name, the table name is not concatenated.  Table \ref{tbl:scn} shows SCN examples; the three columns with the same name \textit{id} are distinguished using their SCNs.

 \section{Experiment}

\subsection{Experiment Setup}
\paragraph{Dataset.} Spider dataset \citep{Yu:18} is used to evaluate our proposed system. We use the same data split as \citet{Yu:18}; 206 databases are split into 146 train, 20 dev, and 40 test. All questions for the same database are in the same split; there are 8659 questions for train, 1034 for dev, and 2147 for test. The test set of Spider is not publicly available, so for testing our models are submitted to the data owner. For evaluation, we used exact matching accuracy, with the same definition as defined in \citet{Yu:18}.
\paragraph{Implementation.}
The proposed system is implemented with Tensorflow \citep{tensorflow2015-whitepaper}. Layernorm \citep{layernorm} and dropout \citep{dropout} are applied between layers, with a dropout rate of 0.1. Exponential decay with decay rate 0.8 is applied for every 3 epochs. On each epoch, the trained classifier is evaluated against the validation dataset, and the training stops when the exact match score for the validation dataset is not improved for 20 consequent training epochs. Minibatch size is set to 16; learning rate is set to $4e^{-4}$. Loss is defined as the sum of all classification losses from the slot-filling decoder.

For BERT-based input encoding, we downloaded the publicly available pre-trained model of BERT, \texttt{BERT-Large, Uncased (Whole Word Masking)}, and fine-tuned the model during training. The learning rate is set to $1e^{-5}$, and minibatch size is set to 4.

\subsection{Experimental Results}

Table \ref{tbl:eval} shows the comparisons of our system with several state-of-the-art systems; Evaluation scores for dev and test datasets are retrieved from the Spider leaderboard\footnote{https://yale-lily.github.io/spider}. The performance of the proposed system is compared with grammar-based systems GrammarSQL \citep{grammarsql}, Global-GNN \citep{gnn} and IRNet \citep{irnet}. Also, we compared the system performance with RCSQL \citep{rcsql}, which so far showed the best performance on the Spider dataset using a sketch-based slot-filling approach.

\begin{table}
\centering
\begin{tabular}{|l|c|c|} \hline
\textbf{System}&\textbf{Dev}&\textbf{Test} \\ \hline
RCSQL & 28.5\% & 24.3\% \\
GrammarSQL & 34.8\% & 33.8\% \\
IRNet&53.2\%&46.7\% \\ 
Global-GNN&52.7\%&47.4\% \\
IRNet v2(BERT)&63.9\%&55.0\% \\ \hline
\textbf{Ours}&& \\
\textbf{RYANSQL}&43.4\%&- \\
\textbf{RYANSQL(BERT)}&\textbf{66.6\%}&\textbf{58.2\%} \\ \hline
\end{tabular}
\caption{Comparison results with other state-of-the-art systems}
\label{tbl:eval}
\end{table}

As can be observed from the table, the proposed system RYANSQL improves the previous slot filling based system RCSQL by a large margin of 15\%p on the development dataset. With the use of BERT, our system outperforms the current state-of-the-art by 3.2\%p on the hidden test dataset, in terms of exact matching accuracy.

Ablation studies are conducted to further figure out the effect of SPC and the two input manipulation methods.  Since the test dataset is not publicly available, we use the development dataset to run the tests. The results are presented in Table \ref{tbl:abl}. In addition, the ablation study results of \textbf{RYANSQL(BERT)} for each hardness level is presented in Table \ref{tbl:diff}. The definitions of hardness levels are the same as the definitions in \citet{Yu:18}. In the tables, \textbf{Proposed} means our proposed system, while \textbf{-SPC} means the one without Statement Position Code, \textbf{-JTF} means the one without \texttt{JOIN} Table Filtering, and \textbf{-SCN} means the one without Supplemented Column Names. 

\begin{table}
\centering
\begin{tabular}{|l|c|c|} \hline
\multirow{2}{*}{\textbf{Approach}}&\multirow{2}{*}{\textbf{RYANSQL}}&\textbf{RYANSQL} \\
& & \textbf{(BERT)} \\ \hline
\textbf{Proposed} & 43.4\% & 66.6\% \\
- \textbf{SPC} &  38.1\% & 57.3\% \\
- \textbf{JTF} & 41.4\% & 63.9\% \\
- \textbf{SCN} & 37.7\% & 56.1\% \\ \hline
\end{tabular}
\caption{Ablation study results of \textbf{RYANSQL} and \textbf{RYANSQL(BERT)}.}
\label{tbl:abl}
\end{table}

\begin{table}
\centering
\begin{tabular}{|l|c|c|c|c|} \hline
\multirow{2}{*}{\textbf{Approach}} & \multirow{2}{*}{\textbf{Easy}} & \textbf{Med-} & \multirow{2}{*}{\textbf{Hard}} & \textbf{Extra} \\
&&\textbf{ium}&&\textbf{Hard} \\ \hline
\textbf{Proposed}&86.0\%&70.5\%&54.6\%&40.6\% \\ 
\textbf{-SPC}&85.6\%&66.6\%&27.0\%&22.4\% \\ 
\textbf{-JTF}&86.8\%&66.1\%&46.6\%&42.4\% \\ 
\textbf{-SCN}&76.4\%&58.2\%&46.6\%&30.6\% \\ \hline
\end{tabular}
\caption{Ablation study results of \textbf{RYANSQL(BERT)} for each hardness level.}
\label{tbl:diff}
\end{table}

As can be observed from the tables, the use of SPC significantly improves the system performance, especially for \textbf{Hard} and \textbf{Extra Hard} queries. The result suggests that by introducing the SPC the proposed system could effectively handle the nested queries. The \textbf{JTF} feature showed some improvements over \textbf{Medium} and \textbf{Hard} queries, meaning that the \textbf{JTF} feature is effective for handling the statements with multiple tables and clauses.

Finally, the \textbf{SCN} feature showed the most significant performance improvement among the three proposed features. When the \textbf{SCN} feature is used, the system performances of all hardness levels are improved significantly. The evaluation result suggests that our proposed input encoder network architectures do not integrate the table names efficiently during the encoding process. But the evaluation result also shows that the proposed system could successfully integrate the table names into encoding vectors by simply applying the proposed \textbf{SCN} feature, instead of modifying the network architectures.

\subsection{Error Analysis}
We analyzed 345 failed examples of the \textbf{RYANSQL(BERT)} system on the development set. 195 of those examples are analyzed to figure out the main reasons for failure.

The most common cause of failure is column selection failure; 68 out of 195 cases (34.9\%) suffered from the error. In many of those cases, the correct column name is not mentioned in a question; for example, for the question ``What is the airport name for airport `AKO'?", the decoder chooses column \texttt{AirportName} instead of \texttt{AirportCode} as its \texttt{WHERE} clause condition column. As mentioned in \citet{cellvalue}, cell value examples for each column will be helpful to solve this problem.

The second frequent error is table number classification error; 49 out of 195 cases (25.2\%) belong to the category. The decoder occasionally chooses too many tables for the \texttt{FROM} clause, resulting unnecessary table \texttt{JOIN}s. Similarly, 22 out of 195 cases (11.3\%) were due to condition number classification error. Those errors could be resolved by observing and updating the extracted slot values as a whole; our future work will focus on this problem.

The remaining 150 errors were either hard to be classified into one category, and some of them were due to different representations of the same meaning, for example: ``\texttt{SELECT} max(age) \texttt{FROM} Dogs" vs. ``\texttt{SELECT} age \texttt{FROM} Dogs \texttt{ORDER BY} age \texttt{DESC} \texttt{LIMIT} 1".

\section{Conclusion}
In this paper, we proposed a sketch-based slot filling algorithm for complex, cross-domain Text-to-SQL problem. A detailed sketch for complex \texttt{SELECT} statement prediction is proposed, along with the Statement Position Code to handle nested queries. Also, two input manipulation methods are proposed to enhance the overall system performance further. Our proposed system achieved the state-of-the-art performance on the challenging Spider benchmark dataset.

The error analysis suggests that we should update slot values based on other slots' prediction results. Our future work will focus on this slot value updating problem.

\bibliography{acl2020_sql2txt}

\begin{thebibliography}{25}
\expandafter\ifx\csname natexlab\endcsname\relax\def\natexlab#1{#1}\fi

\bibitem[{Abadi et~al.(2015)Abadi, Agarwal, Barham, Brevdo, Chen, Citro,
  Corrado, Davis, Dean, Devin, Ghemawat, Goodfellow, Harp, Irving, Isard, Jia,
  Jozefowicz, Kaiser, Kudlur, Levenberg, Man\'{e}, Monga, Moore, Murray, Olah,
  Schuster, Shlens, Steiner, Sutskever, Talwar, Tucker, Vanhoucke, Vasudevan,
  Vi\'{e}gas, Vinyals, Warden, Wattenberg, Wicke, Yu, and
  Zheng}]{tensorflow2015-whitepaper}
Mart\'{\i}n Abadi, Ashish Agarwal, Paul Barham, Eugene Brevdo, Zhifeng Chen,
  Craig Citro, Greg~S. Corrado, Andy Davis, Jeffrey Dean, Matthieu Devin,
  Sanjay Ghemawat, Ian Goodfellow, Andrew Harp, Geoffrey Irving, Michael Isard,
  Yangqing Jia, Rafal Jozefowicz, Lukasz Kaiser, Manjunath Kudlur, Josh
  Levenberg, Dandelion Man\'{e}, Rajat Monga, Sherry Moore, Derek Murray, Chris
  Olah, Mike Schuster, Jonathon Shlens, Benoit Steiner, Ilya Sutskever, Kunal
  Talwar, Paul Tucker, Vincent Vanhoucke, Vijay Vasudevan, Fernanda Vi\'{e}gas,
  Oriol Vinyals, Pete Warden, Martin Wattenberg, Martin Wicke, Yuan Yu, and
  Xiaoqiang Zheng. 2015.
\newblock \href {https://www.tensorflow.org/} {{TensorFlow}: Large-scale
  machine learning on heterogeneous systems}.
\newblock Software available from tensorflow.org.

\bibitem[{Ba et~al.(2016)Ba, Kiros, and Hinton}]{layernorm}
Jimmy~Lei Ba, Jamie~Ryan Kiros, and Geoffrey~E. Hinton. 2016.
\newblock Layer normalization.
\newblock \emph{Computing Research Repository}, arXiv:1607.06450.

\bibitem[{Bogin et~al.(2019)Bogin, Gardner, and Berant}]{gnn}
Ben Bogin, Matt Gardner, and Jonathan Berant. 2019.
\newblock Global reasoning over database structures for text-to-sql parsing.
\newblock In \emph{Proceedings of the 2019 Conference on Empirical Methods in
  Natural Language Processing and the 9th International Joint Conference on
  Natural Language Processing (EMNLP-IJCNLP)}, pages 3650--3655.

\bibitem[{Devlin et~al.(2019)Devlin, Chang, Lee, and Toutanova}]{Devlin:19}
Jacob Devlin, Ming-Wei Chang, Kenton Lee, and Kristina Toutanova. 2019.
\newblock Bert: Pre-training of deep bidirectional transformers for language
  understanding.
\newblock In \emph{Proceedings of the 2019 Conference of the North American
  Chapter of the Association for Computational Linguistics: Human Language
  Technologies}, volume~1, pages 4171--4186.

\bibitem[{Dong and Lapata(2016)}]{dong:16}
Li~Dong and Mirella Lapata. 2016.
\newblock Language to logical form with neural attention.
\newblock In \emph{Proceedings of the 54th Annual Meeting of the Association
  for Computational Linguistics}, pages 33--43.

\bibitem[{Dong and Lapata(2018)}]{dong:18}
Li~Dong and Mirella Lapata. 2018.
\newblock Coarse-to-fine decoding for neural semantic parsing.
\newblock In \emph{Proceedings of the 56th Annual Meeting of the Association
  for Computational Linguistics}, pages 731--742.

\bibitem[{Guo et~al.(2019)Guo, Zhan, Gao, Xiao, Lou, Liu, and Zhang}]{irnet}
Jiaqi Guo, Zecheng Zhan, Yan Gao, Yan Xiao, Jian-Guang Lou, Ting Liu, and
  Dongmei Zhang. 2019.
\newblock Towards complex text-to-sql in cross-domain database with
  intermediate representation.
\newblock \emph{Computing Research Repository}, arXiv:1905.082057.

\bibitem[{He et~al.(2019)He, Mao, Chakrabarti, and Chen}]{xsql}
Pengcheng He, Yi~Mao, Kaushik Chakrabarti, and Weizhu Chen. 2019.
\newblock X-sql: reinforce schema representation with context.
\newblock \emph{Computing Research Repository}, arXiv:1908.08113.

\bibitem[{Hu et~al.(2018)Hu, Peng, Huang, Qiu, Wei, and Zhou}]{mnemonic}
Minghao Hu, Yuxing Peng, Zhen Huang, Xipeng Qiu, Furu Wei, and Ming Zhou. 2018.
\newblock Reinforced mnemonic reader for machine reading comprehension.
\newblock In \emph{Proceedings of the 27th International Joint Conference on
  Artificial Intelligence}, pages 4099--4106.

\bibitem[{Hwang et~al.(2019)Hwang, Yim, Park, and Seo}]{sqlova}
Wonseok Hwang, Jinyeong Yim, Seunghyun Park, and Minjoon Seo. 2019.
\newblock A comprehensive exploration on wikisql with table-aware word
  contextualization.
\newblock \emph{Computing Research Repository}, arXiv:1902.01069.

\bibitem[{Lee(2019)}]{rcsql}
Dongjun Lee. 2019.
\newblock Clause-wise and recursive decoding for complex and cross-domain
  text-to-sql generation.
\newblock In \emph{Proceedings of the 2019 Conference on Empirical Methods in
  Natural Language Processing and the 9th International Joint Conference on
  Natural Language Processing (EMNLP-IJCNLP)}, pages 6047--6053.

\bibitem[{Lin et~al.(2019)Lin, Bogin, Neumann, Berant, and
  Gardner}]{grammarsql}
Kevin Lin, Ben Bogin, Mark Neumann, Jonathan Berant, and Matt Gardner. 2019.
\newblock Grammar-based neural text-to-sql generation.
\newblock \emph{Computing Research Repository}, arXiv:1905.13326.

\bibitem[{Mou et~al.(2016)Mou, Men, Li, Xu, Zhang, Yan, and
  Jin}]{heuristicmatching}
Lili Mou, Rui Men, Ge~Li, Yan Xu, Lu~Zhang, Rui Yan, and Zhi Jin. 2016.
\newblock Natural language inference by tree-based convolution and heuristic
  matching.
\newblock In \emph{Proceedings of the 54th Annual Meeting of the Association
  for Computational Linguistics}, pages 130--136.

\bibitem[{Pennington et~al.(2014)Pennington, Socher, and
  Manning}]{pennington14}
Jeffrey Pennington, Richard Socher, and Christopher~D. Manning. 2014.
\newblock \href {http://www.aclweb.org/anthology/D14-1162} {Glove: Global
  vectors for word representation}.
\newblock In \emph{Empirical Methods in Natural Language Processing (EMNLP)},
  pages 1532--1543.

\bibitem[{Shi et~al.(2018)Shi, Tatwawadi, Chakrabarti, Mao, Polozov, and
  Chen}]{incsql}
Tianze Shi, Kedar Tatwawadi, Kaushik Chakrabarti, Yi~Mao, Oleksandr Polozov,
  and Weizhu Chen. 2018.
\newblock Incsql: Training incremental text-to-sql parsers with
  non-deterministic oracles.
\newblock \emph{Computing Research Repository}, arXiv:1809.05054.

\bibitem[{Srivastava et~al.(2014)Srivastava, Hinton, Krizhevsky, Sutskever, and
  Salakhutdinov}]{dropout}
Nitish Srivastava, Geoffrey Hinton, Alex Krizhevsky, Ilya Sutskever, and Ruslan
  Salakhutdinov. 2014.
\newblock Dropout: a simple way to prevent neural networks from overfitting.
\newblock \emph{The Journal of Machine Learning Research}, 15(1):1929--1958.

\bibitem[{Srivastava et~al.(2015)Srivastava, Greff, and Schmidhuber}]{highway}
Rupesh~K. Srivastava, Klaus Greff, and Jürgen Schmidhuber. 2015.
\newblock Training very deep networks.
\newblock \emph{Advances in neural information processing systems.}, pages
  2377--2385.

\bibitem[{Vaswani et~al.(2017)Vaswani, Shazeer, Parmar, Uszkoreit, Jones,
  Gomez, Łukasz Kaiser, and Polosukhin}]{transformer}
Ashish Vaswani, Noam Shazeer, Niki Parmar, Jakob Uszkoreit, Llion Jones,
  Aidan~N. Gomez, Łukasz Kaiser, and Illia Polosukhin. 2017.
\newblock Attention is all you need.
\newblock \emph{Advances in neural information processing systems.},
  30:5998--6008.

\bibitem[{Vinyals et~al.(2015)Vinyals, Fortunato, and Jaitly}]{pointer}
Oriol Vinyals, Meire Fortunato, and Navdeep Jaitly. 2015.
\newblock Pointer networks.
\newblock \emph{Advances in Neural Information Processing Systems}, pages
  2692--2700.

\bibitem[{Xu et~al.(2017)Xu, Liu, and Song}]{sqlnet}
Xiaojun Xu, Chang Liu, and Dawn Song. 2017.
\newblock Sqlnet: Generating structured queries from natural language without
  reinforcement learning.
\newblock \emph{Computing Research Repository}, arXiv:1711.04436.

\bibitem[{Yavuz et~al.(2018)Yavuz, Gur, Su, and Yan}]{cellvalue}
Semih Yavuz, Izzeddin Gur, Yu~Su, and Xifeng Yan. 2018.
\newblock What it takes to achieve 100\% condition accuracy on wikisql.
\newblock \emph{Proceedings of the 2018 Conference on Empirical Methods in
  Natural Language Processing}, pages 1702--1711.

\bibitem[{Yoon et~al.(2018)Yoon, Lee, and Lee}]{Yoon18}
Deunsol Yoon, Dongbok Lee, and Sangkeun Lee. 2018.
\newblock Dynamic self-attention : Computing attention over words dynamically
  for sentence embedding.
\newblock \emph{Computing Research Repository}, arXiv:1808.073837.

\bibitem[{Yu et~al.(2018{\natexlab{a}})Yu, Li, Zhang, Zhang, and
  Radev}]{typesql}
Tao Yu, Zifan Li, Zilin Zhang, Rui Zhang, and Dragomir Radev.
  2018{\natexlab{a}}.
\newblock Typesql: Knowledge-based type-aware neural text-to-sql generation.
\newblock In \emph{Proceedings of the 2018 Conference of the North American
  Chapter of the Association for Computational Linguistics: Human Language
  Technologies}, pages 588--594.

\bibitem[{Yu et~al.(2018{\natexlab{b}})Yu, Zhang, Yang, Yasunaga, Wang, Li, Ma,
  Li, Yao, Roman, Zhang, and Radev}]{Yu:18}
Tao Yu, Rui Zhang, Kai Yang, Michihiro Yasunaga, Dongxu Wang, Zifan Li, James
  Ma, Irene Li, Qingning Yao, Shanelle Roman, Zilin Zhang, and Dragomir Radev.
  2018{\natexlab{b}}.
\newblock Spider: A large-scale human-labeled dataset for complex and
  cross-domain semantic parsing and text-to-sql task.
\newblock In \emph{Proceedings of the 2018 Conference on Empirical Methods in
  Natural Language Processing}, pages 3911--3921.

\bibitem[{Zhong et~al.(2017)Zhong, Xiong, and Socher}]{zhong:17}
Victor Zhong, Caiming Xiong, and Richard Socher. 2017.
\newblock Seq2sql: Generating structured queries from natural language using
  reinforcement learning.
\newblock \emph{Computing Research Repository}, arXiv:1709.00103.

\end{thebibliography}
\bibliographystyle{acl_natbib}

\end{document}